\documentclass[12pt, nosubfloats, final]{l4dc2023}

\title[Improving Gradient Computation for Differentiable Physics Simulation with Contacts]{Improving Gradient Computation for Differentiable Physics Simulation with Contacts}
\usepackage{times}
\usepackage{graphicx}
\usepackage{subfigure}
\usepackage{booktabs} 
\usepackage{multirow}
\usepackage{wrapfig}
\usepackage{soul}
\usepackage{xurl}

\usepackage{amssymb}
\usepackage{pifont}
\newcommand{\cmark}{\ding{51}}%
\newcommand{\xmark}{\ding{55}}%

\author{%
 \Name{Yaofeng Desmond Zhong} \Email{yaofeng.zhong@siemens.com}\\
 \addr Siemens Corporation, Technology
 \AND
 \Name{Jiequn Han} \Email{jiequnhan@gmail.com}\\
 \addr Flatiron Institute
 \AND
 \Name{Biswadip Dey} \Email{biswadip.dey@siemens.com}\\
 \addr Siemens Corporation, Technology
 \AND
 \Name{Georgia Olympia Brikis} \Email{georgia.brikis@siemens.com}\\
 \addr Siemens Corporation, Technology
}

\begin{document}

\maketitle

\begin{abstract}%
Differentiable simulation enables gradients to be back-propagated through physics simulations. In this way, one can learn the dynamics and properties of a physics system by gradient-based optimization or embed the whole differentiable simulation as a layer in a deep learning model for downstream tasks, such as planning and control. However, differentiable simulation at its current stage is not perfect and might provide wrong gradients that deteriorate its performance in learning tasks. In this paper, we study differentiable rigid-body simulation with contacts. We find that existing differentiable simulation methods provide inaccurate gradients when the contact normal direction is not fixed - a general situation when the contacts are between two moving objects. We propose to improve gradient computation by continuous collision detection and leverage the time-of-impact (TOI) to calculate the post-collision velocities. We demonstrate our proposed method, referred to as TOI-Velocity, on two optimal control problems. We show that with TOI-Velocity, we are able to learn an optimal control sequence that matches the analytical solution, while without TOI-Velocity, existing differentiable simulation methods fail to do so. 
\end{abstract}

\begin{keywords}%
  differentiable simulation, rigid-body simulation, collision and contacts, optimal control
\end{keywords}

\section{Introduction}
With rapid advances and development of machine learning and automatic differentiation tools, a family of techniques has emerged to make physics simulation end-to-end differentiable \citep{liang2020differentiable}. These differentiable physics simulators make it easy to use gradient-based methods for learning and control tasks, such as system identification \citep{zhong2021extending, le2021differentiable, pmlr-v120-song20a}, learning to slide unknown objects \citep{song2020learning}, shape optimization \citep{strecke2021diffsdfsim, Xu_RSS_21} and grasp synthesis \citep{turpin2022grasp}. These applications demonstrate the potential of differentiable simulations in solving control and design problems that are hard to solve with traditional tools.
Compared to black-box neural network counterparts, differentiable simulations utilize physical models to provide more reliable gradient information and better interpretability, which benefits various learning tasks involving physics simulations. One important and popular category of differentiable simulation investigates rigid-body simulation with collisions and contacts. However, current methods might compute gradients incorrectly, providing useless or even harmful signals for learning and optimization tasks. In the present study, we directly identify why wrong gradients occur in the original optimization problem and propose a novel technique to improve gradient computation. Our results on two optimal control examples clearly show the advantage of our proposed method. It is worth noting that another line of research in recent literature has attempted to address the challenge by modifying the optimization problem by incorporating randomness into the objective function \citep{suh2022differentiable, suh2022bundled, lidec2022augmenting} or implementing a smooth critic \citep{xu2022accelerated}.

\section{Preliminaries}
\subsection{Differentiable Simulation with Contacts}
\label{sec:diff_sim_w_contact}
In this section, we provide a brief overview of the different types of differentiable contact models. We classify these methods into the following two categories.

\textbf{Velocity-impulse-based methods} treat contact events as instantaneous velocity changes. There are many ways to solve these velocity impulses. For frictional contacts, the problem can be formulated as a nonlinear complementarity problem (NCP) \citep{howell2022dojo}. Most existing works approximate the NCP by a \emph{linear complementarity problem (LCP)} and apply different techniques to calculate the gradients \citep{de2018end, heiden2021neuralsim, degrave2019differentiable, qiao2021efficient, werling2021fast, du2021diffpd, li2021diffcloth}. Another line of research solves velocity impulses by formulating it as a \emph{convex optimization problem} \citep{todorov2011convex, todorov2012mujoco, todorov2014convex}.  \citet{zhong2021extending} implement a differentiable version of it with CvxpyLayer \citep{agrawal2019differentiable}. If the contact is frictionless, the velocity impulses can be computed directly in a straightforward way, e.g., as done in \cite{chen2021learning}, and we refer to it as the \emph{direct velocity impulse} method.

\textbf{Non-velocity-impulse-based methods} treat contact events in different ways. 
\emph{Compliant models} resolve contact in multiple consecutive time steps and are studied extensively in the context of differentiable simulation \citep{carpentier2018analytical, xu2022accelerated, heiden2021neuralsim, murthy2021gradsim, geilinger2020add, li2020incremental, heiden2021disect, du2021diffpd, warp2022, geilinger2020add}. Besides compliant models, \emph{position-based dynamics (PBD)} \citep{muller2007position, macklin2016xpbd} that directly manipulate positions to resolve contacts can also be easily made differentiable as done in \cite{warp2022, brax2021github, macklin2020primal, liang2019differentiablecloth, qiao2020scalable, yang2020learning, Hu2020DiffTaichi}.

We refer readers to \citet{zhong2022differentiable} for a more detailed overview of these methods. 


\subsection{Time-of-impact - Position}
\cite{Hu2020DiffTaichi} is among the first to study differentiable simulation with contacts from the perspective of velocity impulses. They find that the most straightforward implementation might produce wrong gradients due to time discretization and they propose to use continuous collision detection to find the time-of-impact (TOI) to improve gradient computation. We refer to it as TOI-Position since it leverages TOI to adjust the post-collision position. In this paper, we argue that TOI-Position is not enough to solve all the issues caused by time discretization, especially if the contact normal is not fixed over the optimization iterations. We propose a new technique for velocity-impulse-based methods to improve gradient computation. 

\section{Motivating Problem}
In this section, we revisit an optimal control problem studied by \cite{hu2022solving} and demonstrate that current velocity-impulse-based differentiable simulation methods cannot learn an optimal control input for this problem.

\subsection{Problem Setup}
\begin{wrapfigure}[10]{r}{0.345\textwidth}
    \centering
    \vspace{-50pt}
        \includegraphics[width=0.32\textwidth]{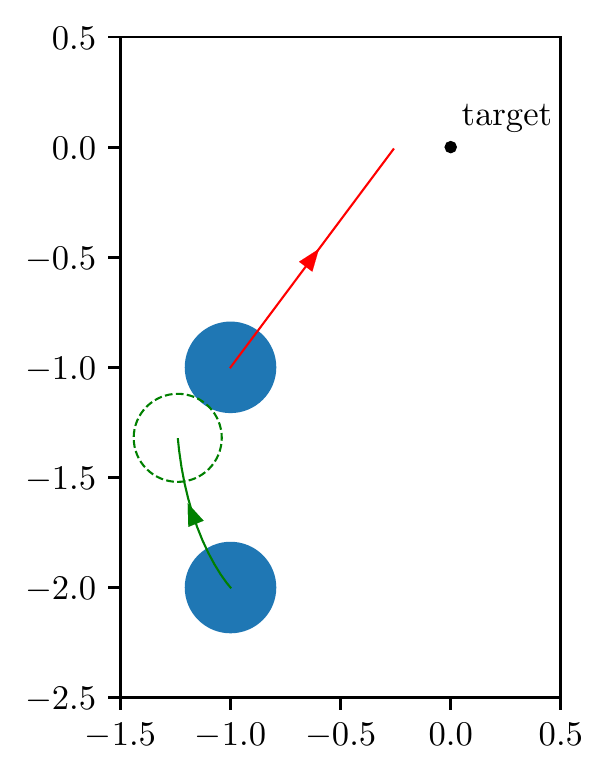}
    \vspace{-10pt}
    \caption{Motivating problem}
    
    \label{fig:two_balls_2_traj}
\end{wrapfigure}
The system is shown in Figure~\ref{fig:two_balls_2_traj}. The two blue circles represent the initial position of two balls, respectively. The pre-collision trajectory of Ball 1 is shown in green, and the post-collision trajectory of Ball 2 is shown in red. The goal is to push Ball 1 to strike Ball 2 so that Ball 2 will reach the target at the end of the simulation. We will formulate it as an optimal control problem later in Section~\ref{sec:exp}. \cite{hu2022solving} uses the hybrid minimum principle (HMP) to obtain an analytical solution to this class of optimal control problems. We will use this analytical solution to evaluate the performance of different methods.

\subsection{Learning with Direct Velocity Impulse}
\begin{figure}[h]
    \centering
        \includegraphics[width=0.6\textwidth]{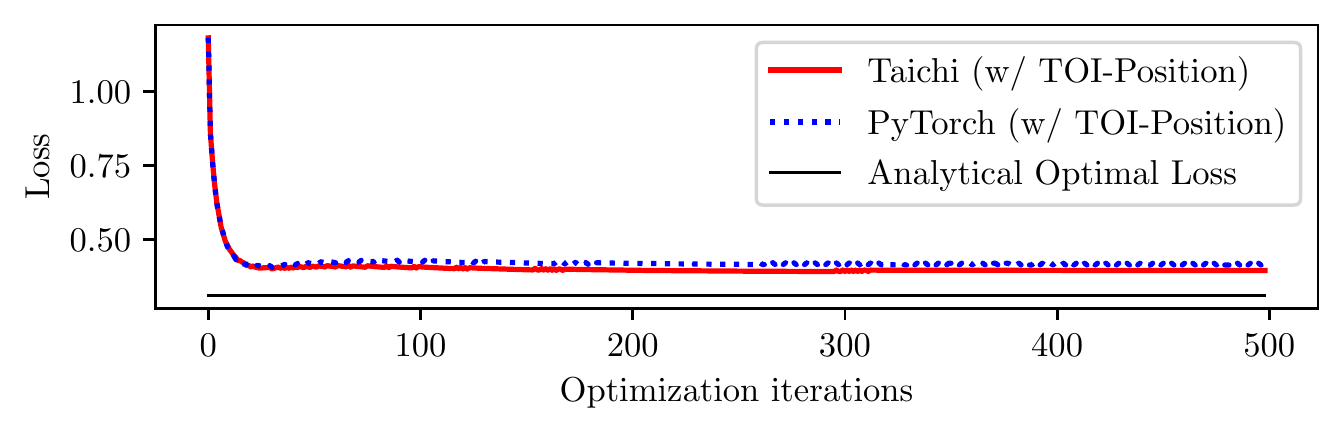}
    \caption{Learning curves of direct velocity impulse method implemented in Taichi and PyTorch.}
    \label{fig:two_balls_2_problematic_learning_curve}
\end{figure}

An optimal control problem can be viewed as a constrained optimization problem where we design the control input at each time step to minimize an objective function. With differentiable simulation, we can compute the gradients of the objective with respect to the control inputs and use gradient-based optimization approaches to learn an optimal control sequence that minimizes the objective/loss function. Figure~\ref{fig:two_balls_2_problematic_learning_curve} shows the learning curves of the direct velocity impulse method implemented in Taichi and PyTorch, along with the analytically obtained optimal value. We observe that both implementations fail to converge to the analytical value even when TOI-Position is used. In Section~\ref{sec:exp}, we will see that the learned control inputs do not match the analytical solution either (Figure~\ref{fig:two_balls_2_loss}). This result highlights the issues present in existing differentiable simulation methods. 

\subsection{A Hint of the Issue}
\begin{figure}[h]
    \centering
        \includegraphics[width=1.0\textwidth]{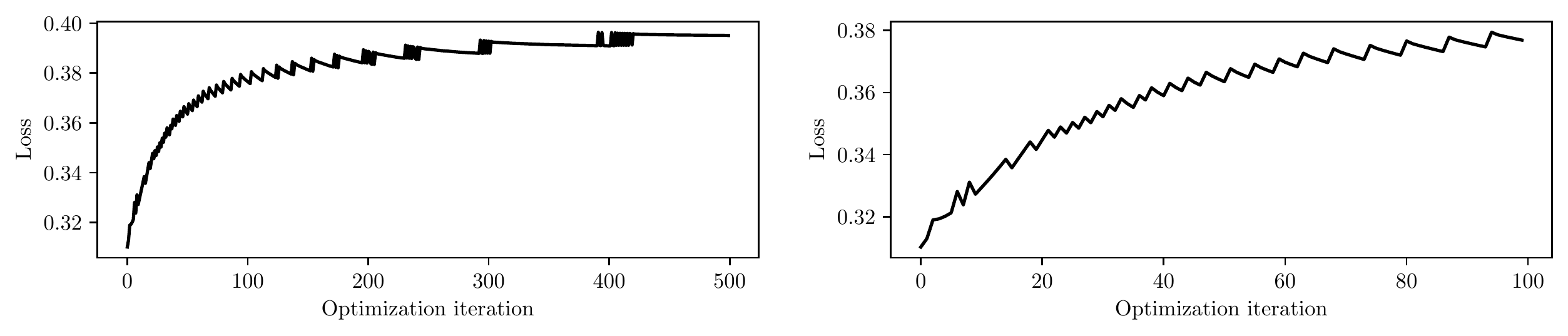}
    \caption{Motivating problem: Learning curve initialized from the analytical optimal solution.}
    \label{fig:two_balls_2_loss_from_u_opt}
\end{figure}

To understand why the learning curve does not converge to the analytically obtained optimal solution, we initiate the learning with the analytical solution. Figure~\ref{fig:two_balls_2_loss_from_u_opt} shows the learning curve over the first 100 iterations. We observe that the loss jumps up in certain iterations, and the learning converges to a solution with a higher loss. This behavior is persistent across different learning rates, which indicates that the gradients computed by the differentiable simulation are incorrect.

\section{Method}
In this section, we first explain why the loss function increases in Figure~\ref{fig:two_balls_2_loss_from_u_opt} and then introduce a new technique, referred to as \emph{TOI-Velocity}, to improve gradient calculation in differentiable simulation. We mainly work with a discrete-time formulation, where the simulation duration $T$ is discretized into $N$ time steps with $\Delta t = T/N$. As the examples in the paper involve two objects moving in the 2D space,  we use $p_n= [p_{1, n}, p_{2, n}]=[p_{x_1, n}, p_{y_1, n}, p_{x_2, n}, p_{y_2, n}]$ to denote the $x$ and $y$ coordinates of the two objects at time step $n$. When we need to distinguish variables from different optimization iterations, we use the superscript $i$ to denote variables in iteration $i$. 
\begin{figure}[h]
\vspace{-1em}
    \centering
    \subfigure[]{
        \centering
        \includegraphics[width=0.29\textwidth]{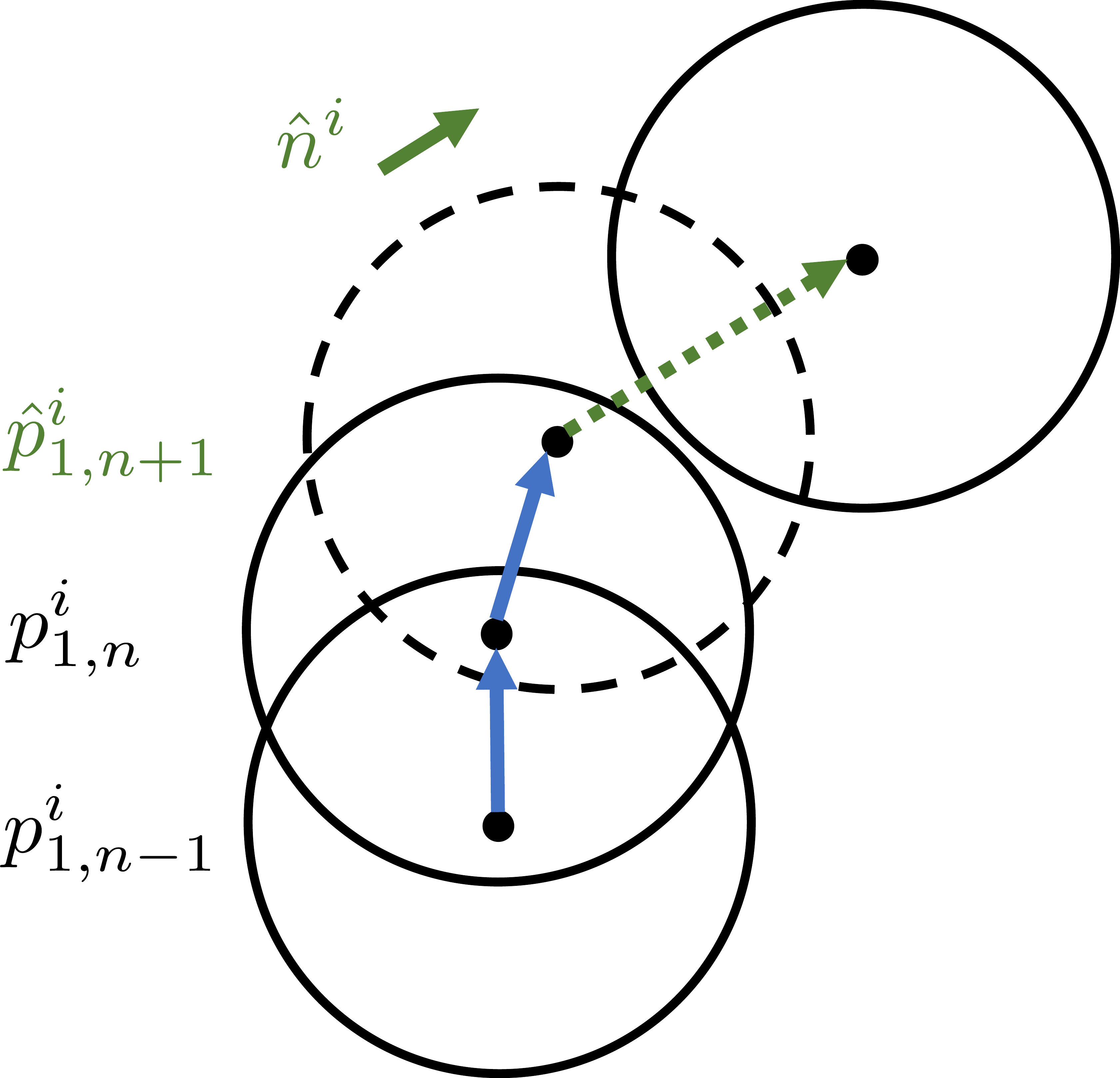}
        \label{fig:cartoon_1}
    }
    \subfigure[]{
        \centering
        \includegraphics[width=0.29\textwidth]{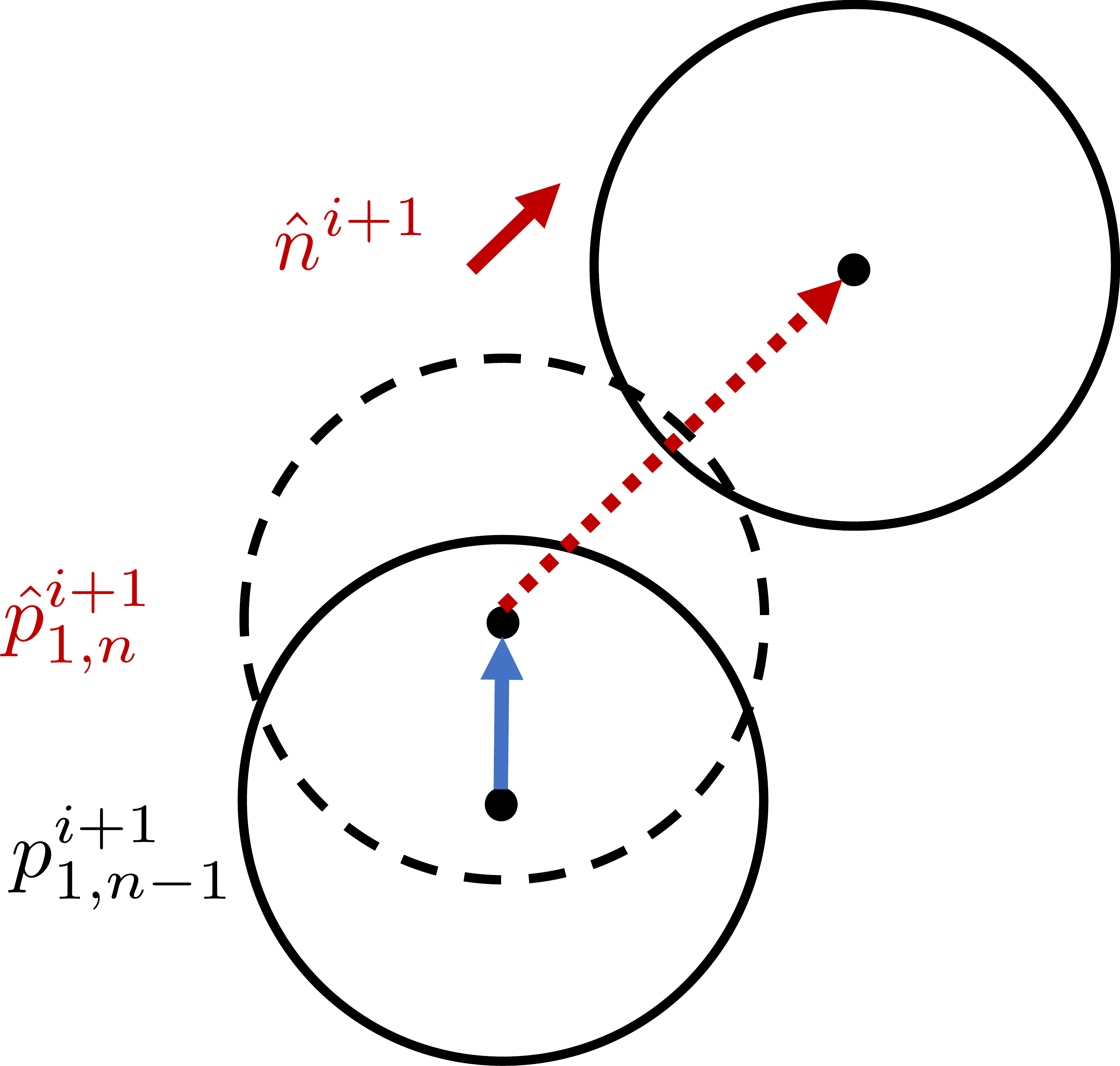}
        \label{fig:cartoon_2}
    }
    \subfigure[]{
        \centering
        \includegraphics[width=0.37\textwidth]{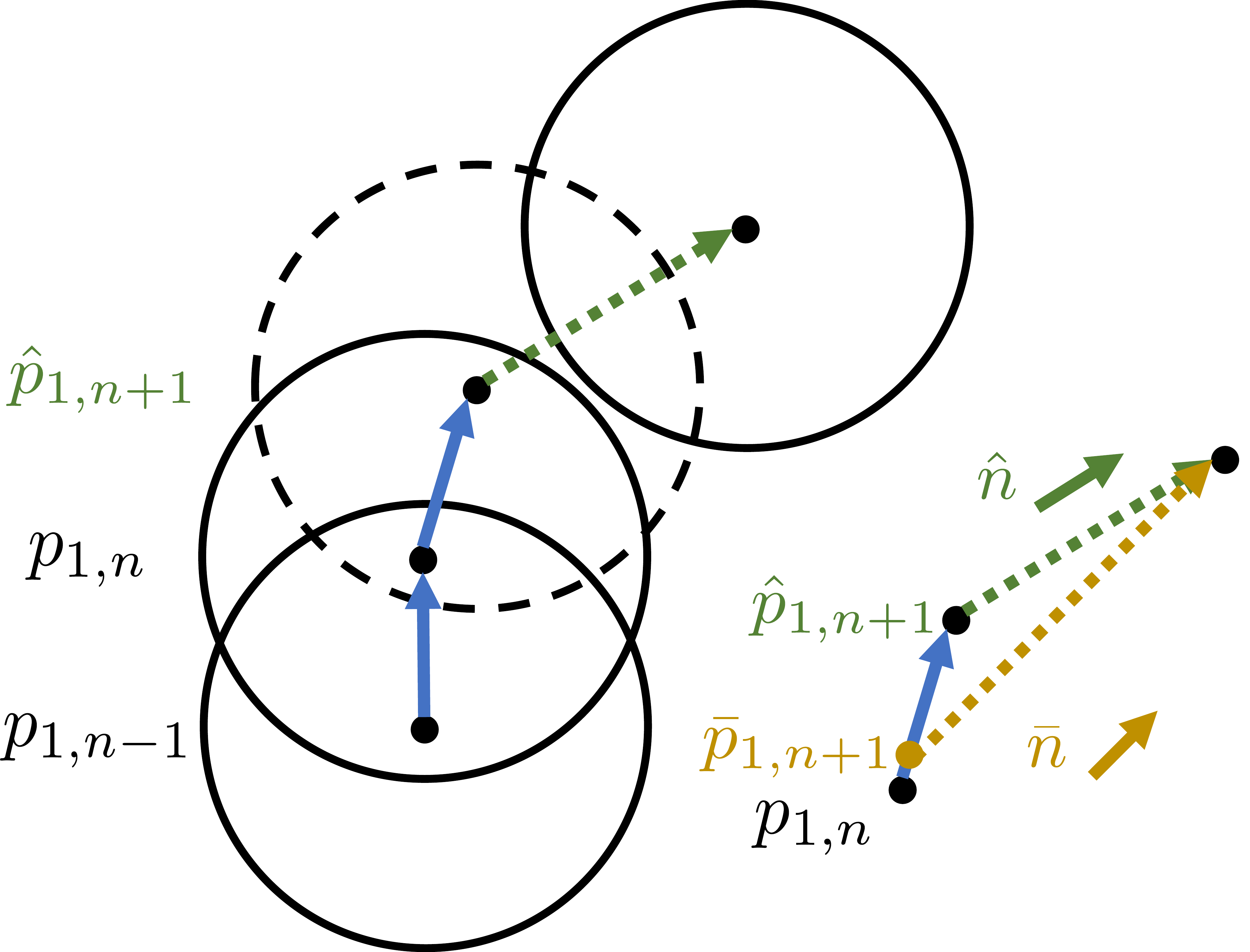}
        \label{fig:cartoon_3}
    }
    \vspace{-0.5em}
    \caption{(a) ball positions in iteration $i$ ; (b) ball positions in iteration $i+1$ ; (c) Difference between penetration direction $\hat{n}$ and collision direction $\Bar{n}$ in an arbitrary iteration. }
    \label{fig:cartoon}
\end{figure}
\subsection{The Reason Behind Loss Increase}
After taking a detailed look, we find that the increase of loss appears when the time step at which the collision happens changes over optimization iterations.  The increase in the loss can be explained using the diagrams in Figure~\ref{fig:cartoon}. In Figure~\ref{fig:cartoon_1}, $p_{1, n-1}^i$ and $p_{1, n}^i$ show the position of the Ball 1 at time step $n-1$ and $n$ in iteration $i$, respectively. $\hat{p}_{1, n+1}^i$ denotes the penetrated position of Ball 1, which is an intermediate variable used to resolve the collision (explained below). The direction of the post-collision velocity of Ball 2 is determined by the penetration direction, indicated by the green arrow. Assume that after a gradient update, the position of the balls changes to the one shown in Figure~\ref{fig:cartoon_2}, where the collision happens in time step $n$ instead of time step $n+1$. Now the direction of the post-collision velocity of Ball 2 is determined by the penetration direction indicated by the red arrow. As the change of penetration direction is not continuous, the change in the post-collision velocity of Ball 2 is also not continuous. Thus, the final position of Ball 2 suffers from a sudden change over these iterations, which could cause an increase in the loss since the terminal cost in the objective depends on the final position of Ball 2. This discontinuity in velocity between consecutive gradient updates is the main reason of the loss increase in Figure~\ref{fig:two_balls_2_loss_from_u_opt}. 

\subsection{Time-of-impact - Velocity}
To improve gradient computation in aforementioned situations where contact normal directions are not fixed, we propose to adjust post-collision velocity by continuous contact detection, and we call the technique TOI-Velocity.  We present TOI-Velocity using the two-ball collision scenario shown in Figure~\ref{fig:cartoon_3}, but the idea applies to collisions among multiple objects with a general class of shapes. 

We assume we are using the symplectic Euler integration scheme, but the idea applies to other integration schemes as well. With symplectic Euler, we have
\begin{align}
    \hat{v}_{n+1} &= v_{n} + u_{n} \Delta t, \\
    \hat{p}_{n+1} &= p_{n} + \hat{v}_{n+1} \Delta t,
\end{align}
where $v_{n} = [v_{1, n}, v_{2, n}]=[v_{x_1, n}, v_{y_1, n}, v_{x_2, n}, v_{y_2, n}]$ is the velocity of two balls at the $n$th time step and $u_{n} = [u_{1, n}, u_{2, n}]=[u_{x_1, n}, u_{y_1, n}, u_{x_2, n}, u_{y_2, n}]$ is the control input to the two balls at the $n$th time step. Variables with a hat ($\hat{\cdot}$) denote intermediate variables before a collision is resolved. If there is no collision detected in this time step, we would have $v_{n+1} = \hat{v}_{n+1}$ and $p_{n+1} = \hat{p}_{n+1}$. If we detect a collision as shown in Figure~\ref{fig:cartoon_3}, we refer to $\hat{p}_{n+1}$ as penetration position and $\hat{v}_{n+1}$ as penetration velocity. The penetration direction is defined as
$
    \hat{n} = (\hat{p}_{2, n+1} - \hat{p}_{1, n+1}) / ||\hat{p}_{2, n+1} - \hat{p}_{1, n+1}||_2
$.
Traditional simulation methods solves the post-collision velocities $v_{n+1}$ using penetration velocity $\hat{v}_{ n+1}$ and penetration direction $\hat{n}$, both of which may be discontinuous across optimization iterations due to time discretization. The TOI-Position proposed by \cite{Hu2020DiffTaichi} computes TOI as the time spent after the penetration appears, i.e.,
\begin{equation}
    TOI = d/\big((\hat{v}_{2, n+1} - \hat{v}_{1, n+1}) \cdot \hat{n}\big).
\end{equation}
where $d = ||\hat{p}_{2, n+1} - \hat{p}_{1, n+1}||_2 - 2r$ is the penetration depth.
In other words, the collision time is estimated by $(\Delta t - TOI)$ after the balls are in position $p_{n}$. Then the post-collision position is adjusted by $p_{n+1} = p_n + \hat{v}_{n+1} \cdot (\Delta t - TOI) + \Tilde{v}_{n+1} \cdot TOI$, where $\Tilde{v}_{n+1}$ is the velocity after resolving collision, which depends on elasticity. However, we argue that adjusting the position only is not enough since the post-collision velocity is also incorrectly estimated. To improve that, we consider the collision position $\Bar{p}_{n+1}$ and velocity $\Bar{v}_{ n+1}$
\begin{align}
    \Bar{v}_{n+1} &= v_{n} + u_{n} (\Delta t - TOI), \\
    \Bar{p}_{n+1} &= p_{n} + \hat{v}_{n+1} (\Delta t - TOI).
\end{align}
The collision direction based on the collision position are then
\begin{equation}
    \Bar{n} = (\Bar{p}_{2, n+1} - \Bar{p}_{1, n+1}) / ||\Bar{p}_{2, n+1} - \Bar{p}_{1, n+1}||_2.
\end{equation}
Figure~\ref{fig:cartoon_3} shows a demonstration of collision position $\Bar{p}_{n+1}$ and collision direction $\Bar{n}$. We propose to use collision velocity $\Bar{v}_{ n+1}$ and collision direction $\Bar{n}$ to solve for post-collision velocities $v_{n+1}$. 
In this way, the post-collision velocities would not suffer from the sudden jump as we observed in Figure~\ref{fig:cartoon_1} and \ref{fig:cartoon_2}. Post-collision position $p_{n+1}$ can be determined accordingly.

\section{Experiments}
\label{sec:exp}
\subsection{Optimal control formulation}
The problem is formulated as an optimal control problem in continuous-time with state jumps: 
\begin{align}
    \underset{u(\cdot)}{\textrm{minimize }}\quad
    &\phi(s(T)) + \int_{0}^{T} L(s(t), u(t)) \mathrm{d} t, \label{eq:prob}\\
    \textrm{subject to }\quad &\dot{s}(t) = f(s(t), u(t)), t \in [0, T]\setminus\cup_{k\in\mathcal{K}}\{\gamma_k\}_{}, \notag \\
    & \psi(s(\gamma_k^-)) = 0, \notag \\
    & s(\gamma_k^+) = g(s(\gamma_k^-)). \notag
\end{align}
Here we use $s=[p, v]$ to denote the positions and velocities of two balls as the state variable. $f$ denotes the state dynamics under external forces $u$; $\cup_{k\in\mathcal{K}}\{\gamma_k\}$ denotes the set of collision instances which is characterized by the collision detection function $\psi$; and $g$ denotes the effect of collisions on the state. We choose the terminal cost to be $\phi(s(T)) = ||p_{2}(T)||_2^2$ to capture our goal of Ball 2 reaching the origin and running cost to be $L(s, u) = \epsilon ||u||_2^2$ to penalize large control inputs. In the experiments, we set $\epsilon=0.01$. See the appendix of \citet{hu2022solving} for the analytical solution of this optimal control problem.
%

To solve the above problem approximately, we discretize the problem into 
\begin{align}
    \label{eqn:discrete_optimal_control}
    \underset{u_0, ..., u_{N-1}}{\textrm{minimize }}\quad
    &\phi(s_N) + \sum_{i=0}^{N-1} L(s_i, u_i) \Delta t,\\
    \textrm{subject to }\quad &s_{i+1} = \texttt{step}(s_i, u_i, \Delta t).
\end{align}
where the \texttt{step} function takes the current state and control as inputs and calculates the next time step state based on dynamics and collisions. 
In our experiments, the simulation time is $T=1s$. The time period is discretized into $N=480$ steps, i.e., $\Delta t = 1/480$. 
As the simulation can be made differentiable, we can differentiate through the \texttt{step} function and solve for the optimal control sequence directly using gradient descent.

\subsection{A Single-Collision Example (Motivating Problem)}
\label{sec:single}
In this section, we demonstrate our proposed techniques using our motivating problem (Figure~\ref{fig:two_balls_2_traj}). We have two balls, of the same size (radius $r = 0.2$) on a plane. The collision between the two balls is frictionless and totally elastic. The initial positions of the balls are $p_{1, 0} = [-1, -2]$ and $p_{2, 0} = [-1, -1]$ and the initial velocities are $v_{1, 0} = v_{2, 0} = [0, 0]$.  We can freely choose control inputs as forces acted on the Ball 1, i.e., $u_{1, n}$. The goal is to push Ball 1 to strike Ball 2 so that Ball 2 would be close to the origin at the end of the simulation. This goal can be formulated as an optimal control problem \eqref{eqn:discrete_optimal_control}, where we set $\epsilon=0.01$. We initiate our control sequence as a constant force $u_{1, n} = [0, 3], n=0, ..., N-1$.

\subsubsection{Performance of existing methods}
We implemented several differentiable simulation methods discussed in Section~\ref{sec:diff_sim_w_contact} - linear complementarity problems (LCP), convex optimization problem (Convex), direct velocity impulse (Direct), compliant model (Compliant) and position-based dynamics (PBD). 

\begin{figure}[t]
    \centering
        \includegraphics[width=\textwidth]{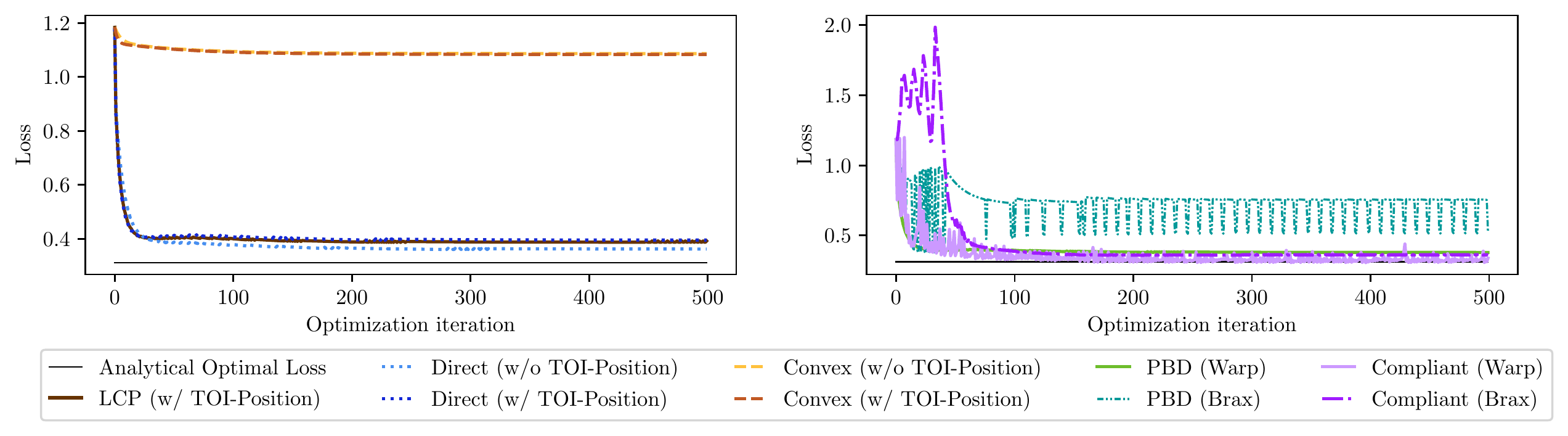}
    \caption{Single-collision: learning curves of different existing differentiable simulation methods. \textbf{Left}: Methods based on velocity impulses. \textbf{Right}: Compliant models and PBD. }
    \label{fig:two_balls_2_loss}
\end{figure}

Figure~\ref{fig:two_balls_2_loss} shows the learning curves of existing methods. The left panel shows all the velocity-impulse-based methods and none of them converges to the analytical optimal loss. The right panel shows non-velocity-impulse-based methods, where many spikes exist in the learning process except PBD implemented in Warp. 

\begin{figure}[h]
    \centering
        \includegraphics[width=\textwidth]{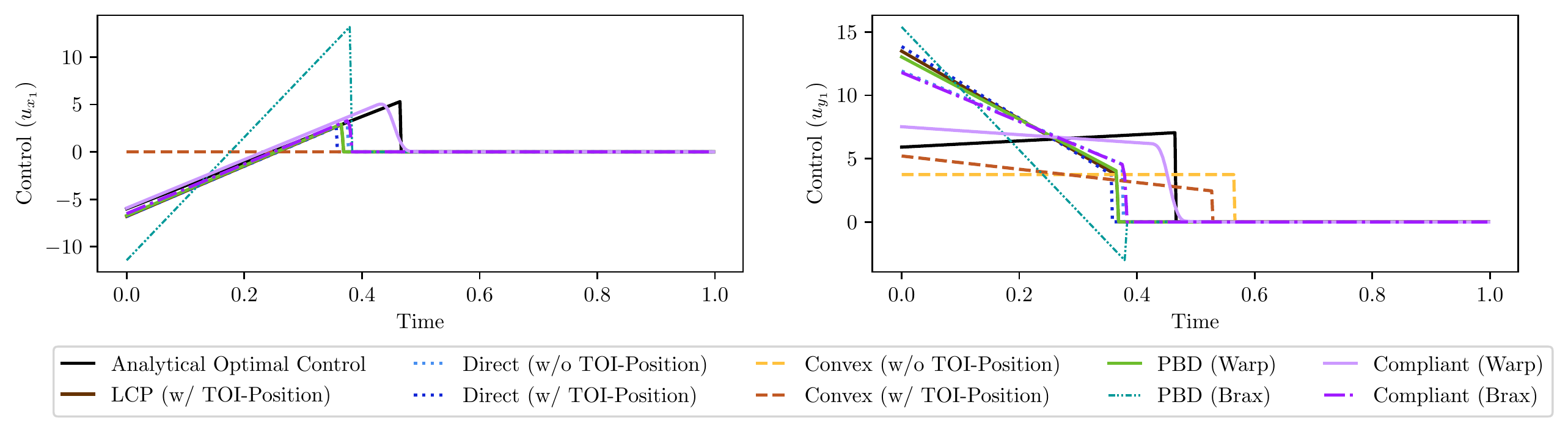}
    \caption{Single-collision: learned control from existing differentiable simulation methods.}
    \label{fig:two_balls_2_ctrls}
\end{figure}

The challenge encountered by existing methods can be further observed from the learned control shown in Figure~\ref{fig:two_balls_2_ctrls}. Even though in the $x$ direction some of the learned control sequences match the shape of the analytical one, in the $y$ direction none of the existing methods is able to learn the correct shape. Specifically, the pre-collision control sequence in the $y$ direction should increase over time while all the learned pre-collision control sequences decrease over time. 

\subsubsection{Performance of TOI-Velocity}
\begin{figure}[h]
    \centering
        \includegraphics[width=0.6\textwidth]{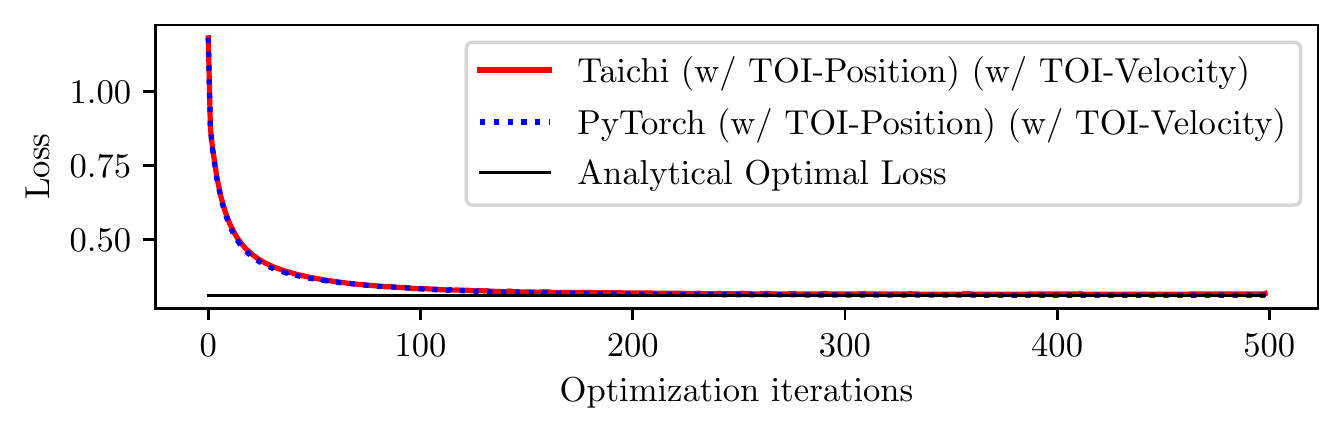}
    \caption{Single-collision: learning curves of our proposed method}
    \label{fig:two_balls_2_proposed_learning_curve}
\end{figure}

\begin{figure}[h]
    \centering
        \includegraphics[width=\textwidth]{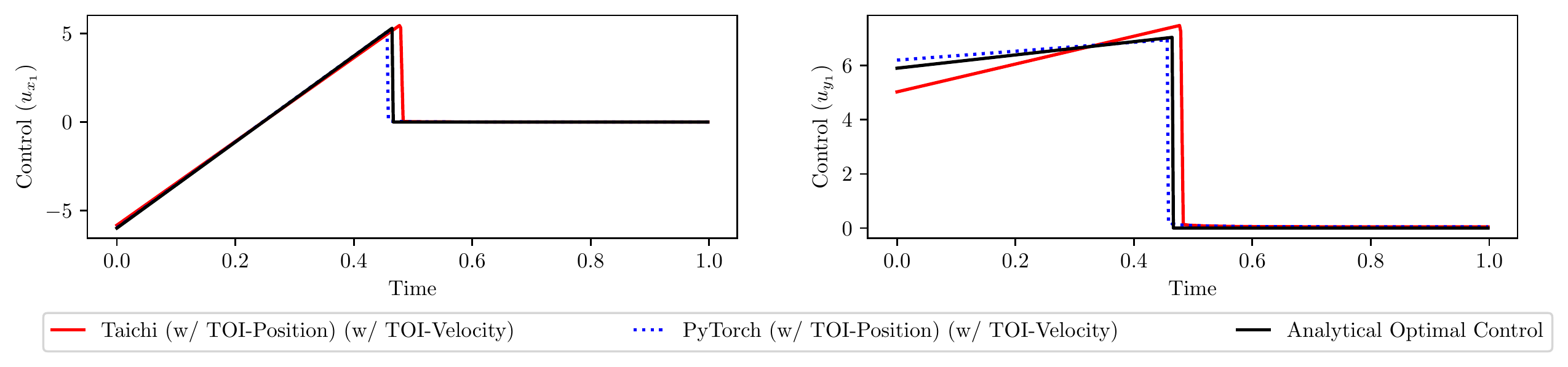}
    \caption{Single-collision: learned control sequence of our proposed method.}
    \label{fig:two_balls_2_proposed_ctrls}
\end{figure}

We implement TOI-Velocity using PyTorch and Taichi. 
The learning curves in Figure~\ref{fig:two_balls_2_proposed_learning_curve} show that both implementations are able to converge to the analytical optimal loss. 
Figure~\ref{fig:two_balls_2_proposed_ctrls} shows the learned control sequences. Both implementations can learn the shape of analytical optimal control and the PyTorch implementation matches the analytical solution better. When comparing Figure~\ref{fig:two_balls_2_proposed_ctrls} with the results of existing methods in Figure~\ref{fig:two_balls_2_ctrls}, we see a clear improvement especially in the $y$ direction, since increasing pre-collision sequences are learned correctly.  

\begin{figure}[b]
\vspace{-1em}
    \centering
    \subfigure[]{
        \centering
        \includegraphics[width=0.48\textwidth]{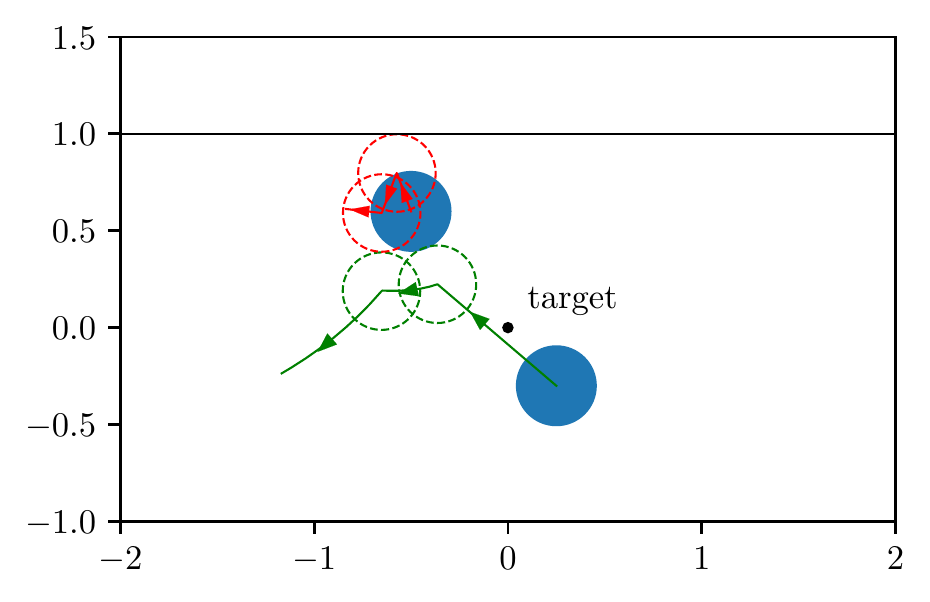}
        \label{fig:two_balls_multiple_initial_traj}
    }
    \subfigure[]{
        \centering
        \includegraphics[width=0.48\textwidth]{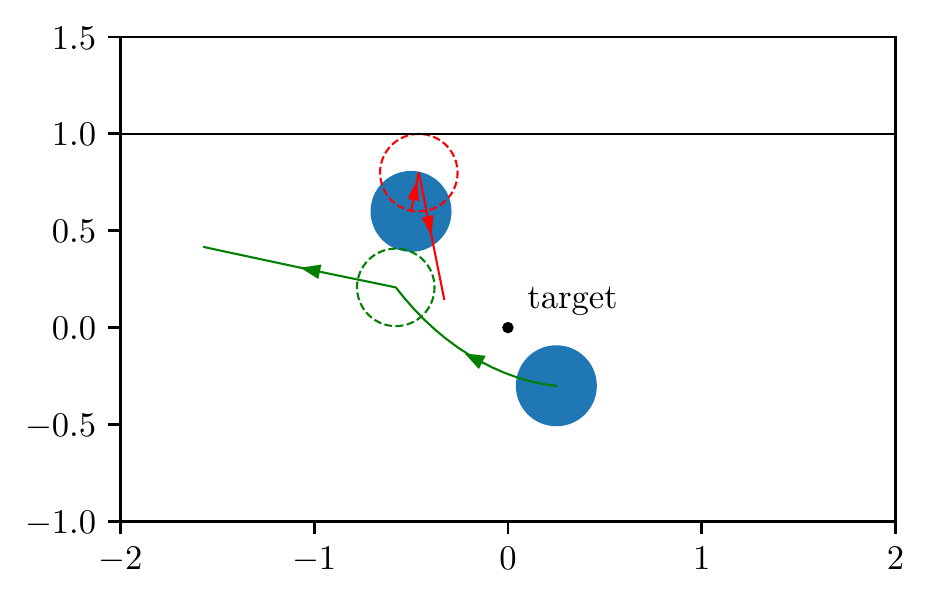}
        \label{fig:two_balls_multiple_optimal_traj}
    }
    \vspace{-0.5em}
    \caption{Multiple-collision: (a) Trajectory before optimization; (b) Analytical optimal trajectory. }
    \label{fig:two_balls_multiple_traj}
\end{figure}
\subsection{A Multiple-Collision Example}
\label{sec:multi}
In this section, we study the system shown in Figure~\ref{fig:two_balls_multiple_traj}, where multiple collisions could happen. 
We have two balls, of the same size (radius $r = 0.2$) on a plane. The initial positions of the balls are $p_{1, 0} = [0.25, -0.3]$ and $p_{2, 0} = [-0.5, 0.6]$ and the initial velocities are $v_{1, 0} = v_{2, 0} = [0, 0]$. There is a wall at location $y=1$. Both the ball-ball collision and ball-wall collision are frictionless and totally elastic.
We can freely choose control inputs as forces acted on the first ball. The goal here is to push Ball 1 to strike Ball 2 so that Ball 2 would be close to the origin at the end of the simulation.
The problem can be formulated as \eqref{eq:prob}.
Figure~\ref{fig:two_balls_multiple_initial_traj} shows the trajectory before optimization where a constant control of $u_{1, n} = [-3.5, 3.0], n=0, ..., N-1$ is applied to Ball 1. This trajectory involves two ball-ball collisions and one ball-wall collision. Figure~\ref{fig:two_balls_multiple_optimal_traj} shows the analytical optimal trajectory, which involves one ball-ball collision and one ball-wall collision. 

\subsubsection{Performance of existing methods}
\begin{figure}[h]
    \centering
        \includegraphics[width=\textwidth]{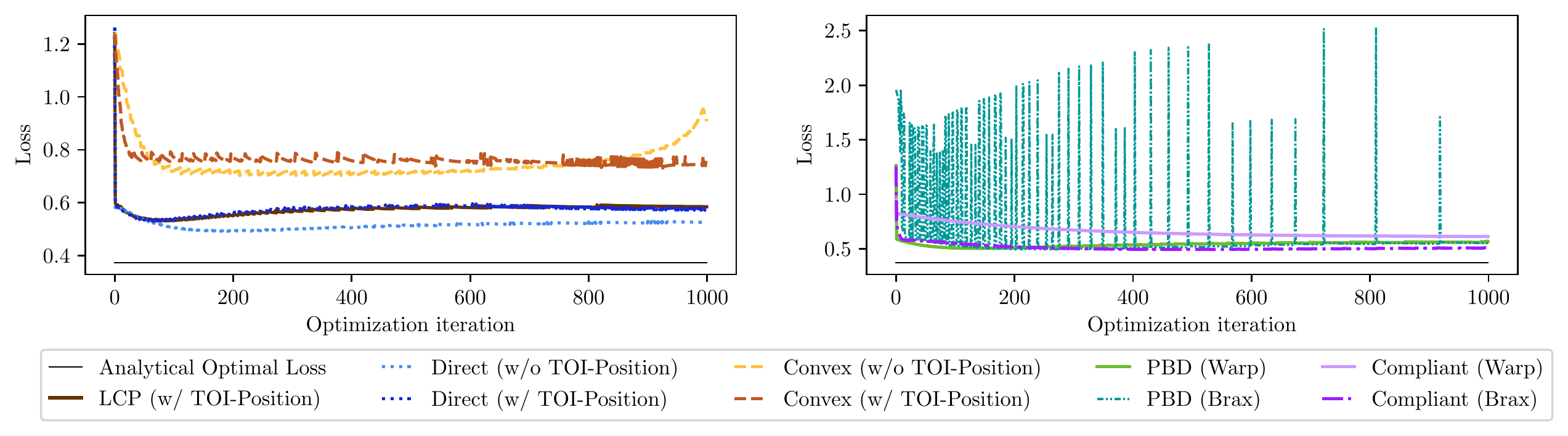}
    \caption{Multiple-collision: learning curves of different existing differentiable simulation methods. \textbf{Left}: Methods based on velocity impulses. \textbf{Right}: Compliant models and PBD.}
    \label{fig:multi_col_loss}
\end{figure}

\begin{figure}[h]
    \centering
        \includegraphics[width=\textwidth]{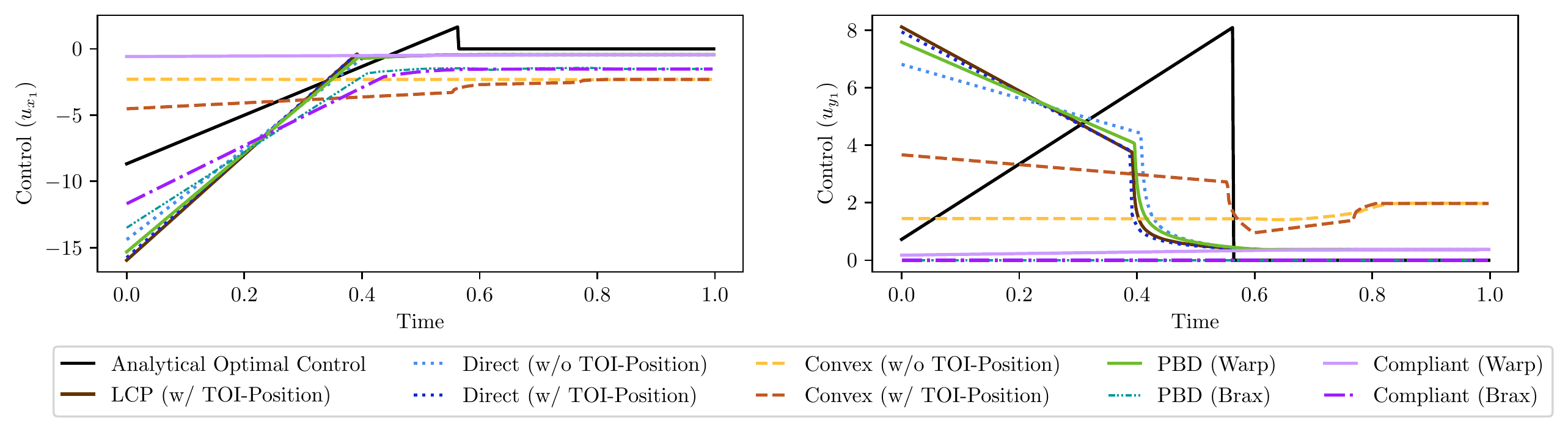}
    \caption{Multiple-collision: learned control sequences of existing methods.}
    \label{fig:multi_col_ctrls}
\end{figure}

Figure~\ref{fig:multi_col_loss} shows the learning curves of existing differentiable simulation methods and none of them converges to the analytical optimal loss. The left panel shows results of velocity-impulse-based methods and for each method, there exists certain time steps where the loss increases over the iterations. This increase of loss is similar to the one observed in Figure~\ref{fig:two_balls_2_loss_from_u_opt}, indicating wrong gradients calculated by the differentiable simulations.  The right panel shows results of non-velocity-impulse-based methods. The learning using the Brax implementation of PBD is unstable as there are many spikes. The other learning curves are smooth but there is a clear gap between the analytical optimal loss and the converged values. Figure~\ref{fig:multi_col_ctrls} shows the control sequences learned by existing differentiable simulation methods and none of them are close to the analytical optimal control. 

\subsubsection{Performance of TOI-Velocity}
\begin{figure}[h]
    \centering
        \includegraphics[width=0.6\textwidth]{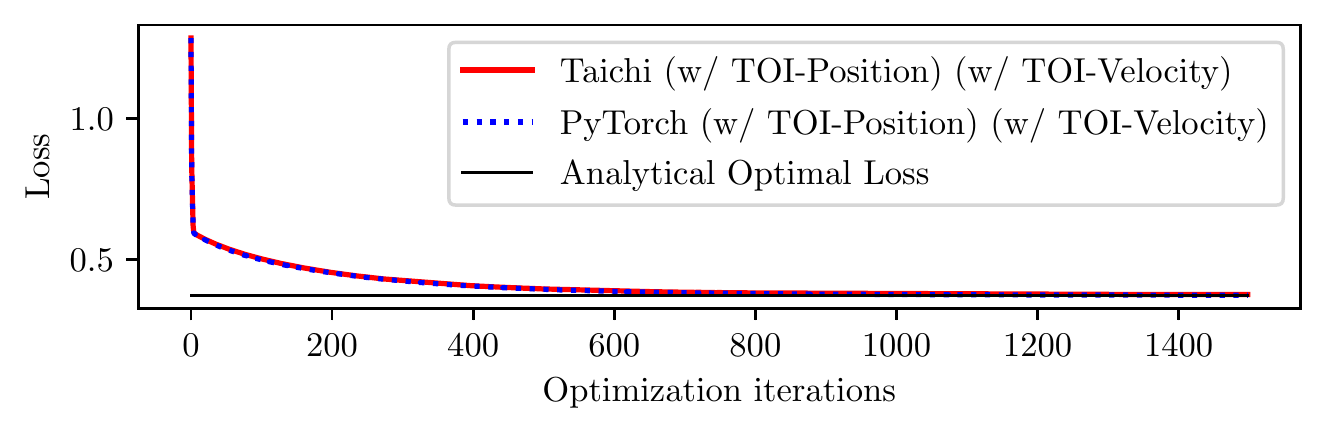}
    \vspace{-15pt}
    \caption{Multiple-collision: learning curves of our proposed method.}
    \label{fig:multi_col_proposed_learning_curve}
\end{figure}

\begin{figure}[h]
    \centering
        \includegraphics[width=1.0\textwidth]{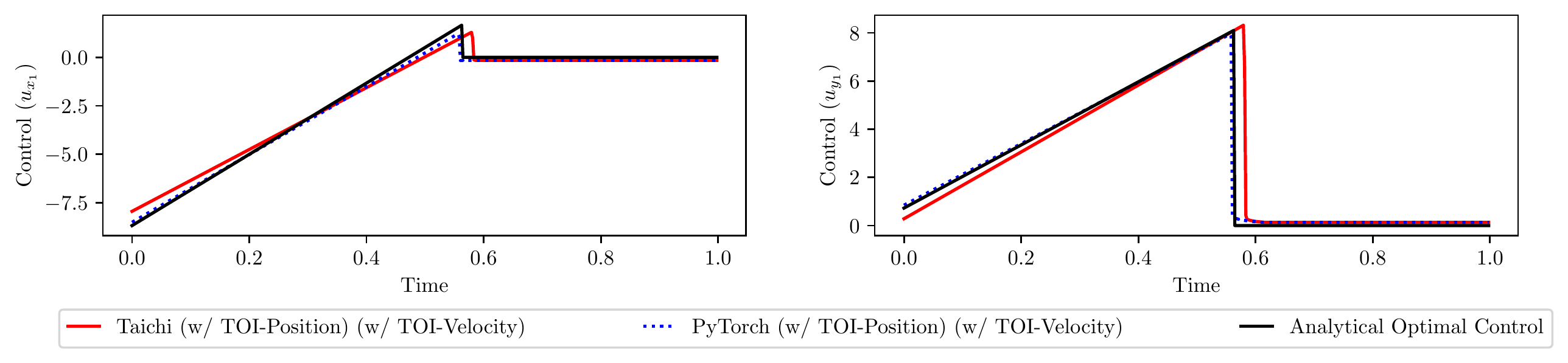}
    \vspace{-18pt}
    \caption{Multiple-collision: learned control of our proposed method.}
    \label{fig:multi_col_proposed_ctrls}
\end{figure}

Applying TOI-Velocity can successfully solve the optimal control problem. Figure~\ref{fig:multi_col_proposed_learning_curve} shows that both Taichi and PyTorch implementation converges to the analytical optimal loss. Figure~\ref{fig:multi_col_proposed_ctrls} shows that the learned control sequences match the analytical optimal control very well. By comparing these results with existing methods, we can conclude that adding the TOI-velocity improves the gradient calculation of velocity-impulse-based differentiable simulation methods. These improved gradients enable us to learn optimal control simply using gradient descent. 

\subsection{Ablation Study}
We have shown that adding TOI-Velocity enables us to successfully learn optimal control, but it is unclear whether TOI-Position is still necessary once TOI-Velocity is applied. Table~\ref{tab:ablation} shows an ablation study on TOI-Position and TOI-Velocity for the two examples in Section~\ref{sec:single} and \ref{sec:multi}. We find that only by applying both TOI-Position and TOI-Velocity can we end up with a loss value close to the analytical optimal loss in both examples. Without TOI-Position, the computed gradients would be wrong, which affects gradient-based learning. 
\begin{table}[h]
\vspace{-5pt}
\caption{Ablation study on TOI-Position and TOI-Velocity}
\label{tab:ablation}
\vskip 0.1in
\centering
\begin{tabular}{c | c | c | c}
    \toprule[1.25pt]
    TOI-Position & TOI-Velocity & single-collision example & multiple-collision example \\
    \midrule[1.0pt]
    \xmark & \xmark & 0.3616 & 0.5261\\
    \cmark & \xmark & 0.3949 & 0.5841\\
    \xmark & \cmark & 0.4797 & 0.6142\\
    \cmark & \cmark & 0.3151 & 0.3785\\
    \midrule[1.0pt]
    \multicolumn{2}{c|}{\textbf{Analytical optimal loss}} & \textbf{0.3115} & \textbf{0.3737}\\
    \bottomrule[1.25pt]
\end{tabular}
\end{table}

\section{Conclusion}
In this paper we propose a novel technique, TOI-Velocity, to reduce discontinuity caused by time discretization in physics simulation. Our proposed method is designed to improve gradient computation in differentiable simulation with contacts. We demonstrate TOI-Velocity in two optimal control examples. Our results show that applying TOI-Velocity together with TOI-Position is the only differentiable simulation implementation that can successfully learn the optimal control in these examples.

\acks{The codebase associated with this work will be released at \url{https://github.com/DesmondZhong/diff_sim_improve_grads}.}

\bibliography{main}

\end{document}